%% file: PaperForReview.tex
\documentclass[10pt,twocolumn,letterpaper]{article}

\usepackage[pagenumbers]{cvpr} % To force page numbers, e.g. for an arXiv version

\usepackage{graphicx}
\usepackage{amsmath}
\usepackage{amssymb}
\usepackage{booktabs}

\usepackage{multirow}
\usepackage{makecell}
\usepackage{graphicx}
\usepackage{enumitem}
\usepackage{comment}
\usepackage{listings}
\usepackage{color, colortbl}

\usepackage{wrapfig}
\usepackage{subcaption} % issues a warning with CVPR/ICCV format
\usepackage{pifont}

\usepackage[accsupp]{axessibility}

\usepackage{bm}
\definecolor{mgreen}{RGB}{1,150,74}
\newcommand\up[1]{\textcolor{mgreen}{$^{\uparrow{#1}}$}}
\newcommand\down[1]{\textcolor{red}{$^{\downarrow{#1}}$}}
\newcommand\blfootnote[1]{%
\begingroup
\renewcommand\thefootnote{}\footnote{#1}%
\addtocounter{footnote}{-1}%
\endgroup
}

\usepackage[pagebackref,breaklinks,colorlinks]{hyperref}

\usepackage[capitalize]{cleveref}
\crefname{section}{Sec.}{Secs.}
\Crefname{section}{Section}{Sections}
\Crefname{table}{Table}{Tables}
\crefname{table}{Tab.}{Tabs.}

\def\cvprPaperID{5343} % *** Enter the CVPR Paper ID here
\def\confName{CVPR}
\def\confYear{2023}

\def\NickName{{GD-MAE}\xspace}
\begin{document}

%%%%%%%%% TITLE - PLEASE UPDATE
\title{\NickName: Generative Decoder for MAE Pre-training on LiDAR Point Clouds}

\author{
Honghui Yang\textsuperscript{\rm 1,3$*$}~~~~
Tong He\textsuperscript{\rm 3$*$}~~~~
Jiaheng Liu\textsuperscript{\rm 3,4}~~~~
Hua Chen\textsuperscript{\rm 5}~~~~
Boxi Wu\textsuperscript{\rm 2}~~~~ \\
Binbin Lin\textsuperscript{\rm 2$\dag$}~~~~
Xiaofei He\textsuperscript{\rm 1}~~~~
Wanli Ouyang\textsuperscript{\rm 3} \\
\textsuperscript{\rm 1}State Key Lab of CAD\&CG, Zhejiang University \\
\textsuperscript{\rm 2}School of Software Technology, Zhejiang University \\
\textsuperscript{\rm 3}Shanghai AI Lab \quad
\textsuperscript{\rm 4}Beihang University \\
\textsuperscript{\rm 5}COMAC Beijing Aircraft Technology Research Institute\\
% {\tt\small yanghonghui@zju.edu.cn}
}
\maketitle

\input{section/abstract}
\input{section/introduction}
\input{section/related_work}

\input{section/method}
\input{section/experiment}
\input{section/conclusion}
\input{section/acknowledgment}

%%%%%%%%% REFERENCES
{\small
\bibliographystyle{ieee_fullname}
\bibliography{egbib}
}

\clearpage

\appendix
\input{section/appendix}

\end{document}

%% file: section/abstract.tex
\begin{abstract}
Despite the tremendous progress of Masked Autoencoders (MAE) in developing vision tasks such as image and video, exploring MAE in large-scale 3D point clouds remains challenging due to the inherent irregularity.
In contrast to previous 3D MAE frameworks, which either design a complex decoder to infer masked information from maintained regions or adopt sophisticated masking strategies, we instead propose a much simpler paradigm.
The core idea is to apply a \textbf{G}enerative \textbf{D}ecoder for MAE (GD-MAE) to automatically merges the surrounding context to restore the masked geometric knowledge in a hierarchical fusion manner.
In doing so, our approach is free from introducing the heuristic design of decoders and enjoys the flexibility of exploring various masking strategies.
The corresponding part costs less than \textbf{12\%} latency compared with conventional methods, while achieving better performance.
We demonstrate the efficacy of the proposed method on several large-scale benchmarks: Waymo, KITTI, and ONCE.
Consistent improvement on downstream detection tasks illustrates strong robustness and generalization capability.
Not only our method reveals state-of-the-art results, but remarkably, we achieve comparable accuracy even with \textbf{20\%} of the labeled data on the Waymo dataset.
Code will be released at \url{https://github.com/Nightmare-n/GD-MAE}.
\blfootnote{$^*$Equal contribution. This work was done when Honghui was an intern at Shanghai Artificial Intelligence Laboratory.}
\blfootnote{$^\dag$Corresponding author}
\end{abstract}

%% file: section/introduction.tex
\section{Introduction}
We have witnessed great success in 3D object detection~\cite{yin2021center,shi2020pvrcnn,yang2022graphrcnn,shi2020part,zheng2021sessd,yang20213dman}, due to the numerous applications in autonomous driving, robotics, and navigation. Despite the impressive performance, most methods count on large amounts of carefully labeled 3D data, which is often of high cost and time-consuming. Such a fully supervised manner hinders the possibility of using massive unlabeled data and can be vulnerable when applied in different scenes.
\begin{figure}[!t]
	\centering
	\includegraphics[width=0.95\columnwidth]{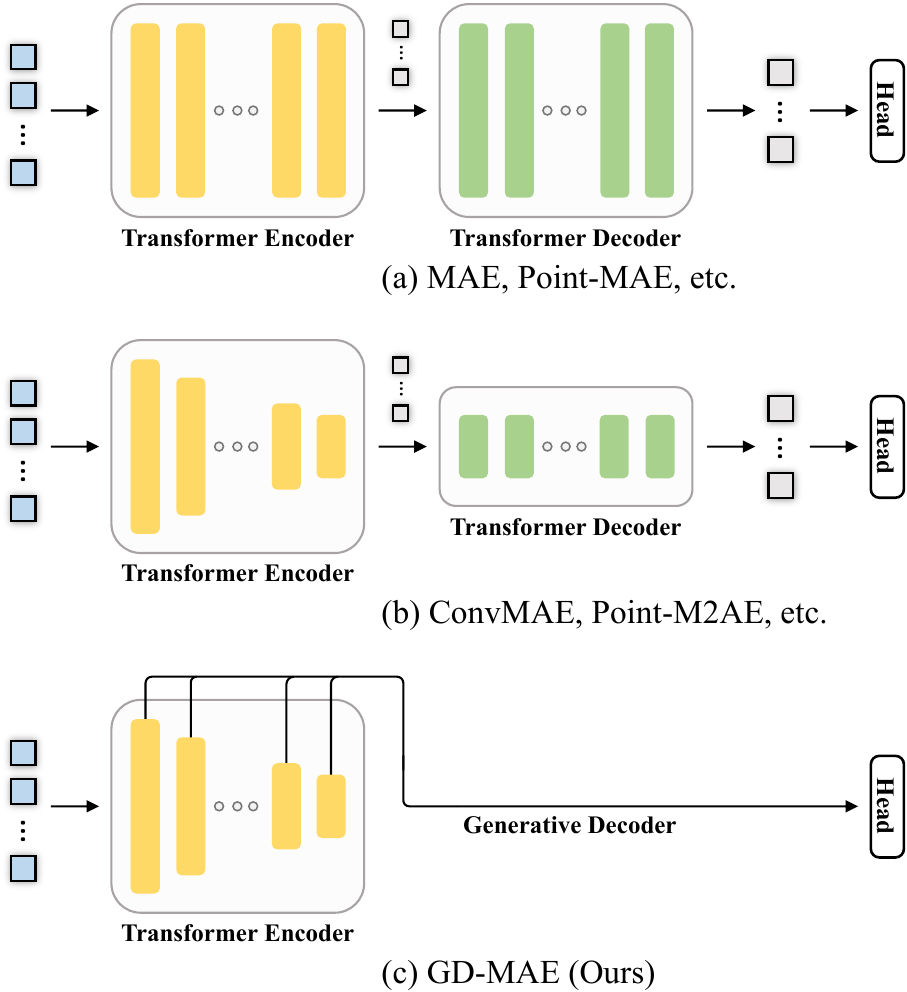}
	\vspace{-0.8em}
	\caption{Comparisons. Previous MAE-style pre-training architectures of (a) single-scale~\cite{he2022mae,pang2022pointmae,hess2022voxelmae} and (b) multi-scale~\cite{zhang2022pointm2ae,gao2022convmae} take as inputs the visible tokens and learnable tokens for decoders. In contrast, (c) the proposed framework avoids such a process.}
	\label{fig:motivation}
	\vspace{-2.0em}
\end{figure}
Mask Autoencoder (MAE)~\cite{he2022mae}, serving as one of the effective ways for pre-training, has demonstrated great potential in learning holistic representations. This is achieved by encouraging the method to learn a semantically consistent understanding of the input beyond low-level statistics. Although MAE-based methods have shown effectiveness in 2D image ~\cite{he2022mae} and video~\cite{tong2022videomae}, how to apply it in large-scale point clouds remains an open problem.

Due to the large variation of the visible extent of objects, learning hierarchical representation is of great significance in 3D supervised learning~\cite{qi2019votenet,shi2019pointrcnn,yan2018second}.
To enable MAE-style pre-training on the hierarchical structure, previous approaches~\cite{zhang2022pointm2ae,gao2022convmae} introduce either complex decoders or elaborate masking strategies to learn robust latent representations.
For example, ConvMAE~\cite{gao2022convmae} adopts a block-wise masking strategy that first obtains a mask for the late stage of the encoder and then progressively upsamples the mask to larger resolutions in early stages to maintain masking consistency.
Point-M2AE~\cite{zhang2022pointm2ae} proposes a hierarchical decoder to gradually incorporate low-level features into learnable tokens for reconstruction.
Meanwhile, it needs a multi-scale masking strategy that backtracks unmasked positions to all preceding scales to ensure coherent visible regions and avoid information leakage.
The minimum size of masking granularity is highly correlated to output tokens of the last stage, which inevitably poses new challenges, especially to objects with small sizes, e.g., pedestrians.

To alleviate the issue, we present a much simpler paradigm dubbed \NickName for pre-training, as shown in Figure~\ref{fig:motivation}.
The key is to use a generative decoder to automatically expand the visible regions to the underlying masked area.
In doing so, it eliminates the need for designing complex decoders, in which masked regions are presented as learnable tokens.
It also allows for the unification of multi-scale features into the same scale, thus enabling flexible masking strategies, e.g., point- and patch-wise masking, while avoiding intricate operations such as backtracking in \cite{zhang2022pointm2ae,gao2022convmae} to keep masking consistency.
Specifically, it consists of the following components:

Firstly, we propose the Sparse Pyramid Transformer (SPT) as the multi-scale encoder.
Following \cite{lang2019pointpillar,shi2022pillarnet,fan2022sst}, SPT takes pillars as input due to the compact and regular representation.
Unlike PointPillars~\cite{lang2019pointpillar} that uses traditional convolutions for feature extraction, we use the sparse convolution~\cite{yan2018second} to downsample the tokens and the sparse transformer~\cite{fan2022sst} to enlarge the receptive field of the visible tokens when deploying extensive masking.

Secondly, we introduce the Generative Decoder (GD) to simplify MAE-style pre-training on multi-scale backbones.
GD consists of a series of transposed convolutions used to upsample multi-scale features and a convolution utilized to expand the visible area, as shown in Figure~\ref{fig:adapter_demo}.
The expanded features are then directly indexed according to the coordinates of the masked tokens for the geometric reconstruction.

Extensive experiments have been conducted on Waymo Open Dataset~\cite{sun2020wod}, KITTI~\cite{geiger2012kitti}, and ONCE~\cite{mao2021once} to verify the efficacy.
On the Waymo dataset, \NickName sets new state-of-the-art detection results compared to previously published methods.

Our contributions are summarized as follows:
\begin{itemize}
\item We introduce a simpler MAE framework that avoids complex decoders and thus simplifies pre-training.
\item The proposed decoder enables flexible masking strategies on LiDAR point clouds, while costing less than 12\% latency compared with conventional methods.
\item Extensive experiments are conducted to verify the effectiveness of the proposed model.
\end{itemize}

%% file: section/related_work.tex
\section{Related Work}
\paragraph{3D Object Detection from Point Clouds.}
With the release of several large-scale LiDAR datasets, there have been many recent networks proposed for 3D object detection~\cite{zheng2020ciassd,zheng2021sessd,he2020sassd,wang2020piilar-od,Yang2018pixor,deng2021voxelrcnn,wu2022sfd}.
VoxelNet~\cite{zhou2018voxelnet} leverages PointNet~\cite{qi2017pointnet} to generate a voxel-wise representation and applies standard convolutions for object detection. 
SECOND~\cite{yan2018second} exploits sparse 3D convolutions to accelerate VoxelNet.
% PointPillars~\cite{lang2019pointpillar} builds on this by further simplifying the voxels to pillars.
Point2Seq~\cite{xue2022point2seq} reformulates the 3D object detection task as decoding words from 3D scenes in an auto-regressive manner.
Due to the quantization errors of voxelization, some methods~\cite{shi2020pointgnn,shi2019pointrcnn,shi2020part,yang2019std,chen2022sasa,wang2022rbgnet,zhang2022iassd} directly operate on raw point clouds for detection.
3DSSD~\cite{yang20203dssd} extends VoteNet~\cite{qi2019votenet} and proposes a hybrid sampling strategy by utilizing both feature and geometry distance for better classification performance.
Sampling and grouping points are generally time-consuming.
Thus, a number of approaches~\cite{ye2020hvnet,miao2021pvgnet,yang2022graphrcnn,mao2021pyramid} take advantage of the efficiency of 3D sparse convolutions while preserving accurate point positions.
% PV-RCNN~\cite{shi2020pvrcnn,shi2021pvrcnnplusplus} extracts point-wise features from 3D voxel backbones to refine the proposals.
FSD~\cite{fan2022fsd} builds a fully sparse 3D object detector to enable efficient long-range detection.
Graph R-CNN~\cite{yang2022graphrcnn} speeds up the RoI pooling in PointRCNN~\cite{shi2019pointrcnn} and introduces a graph-based refinement to achieve better performance.

\begin{figure}[!t]
	\centering
	\includegraphics[width=0.95\columnwidth]{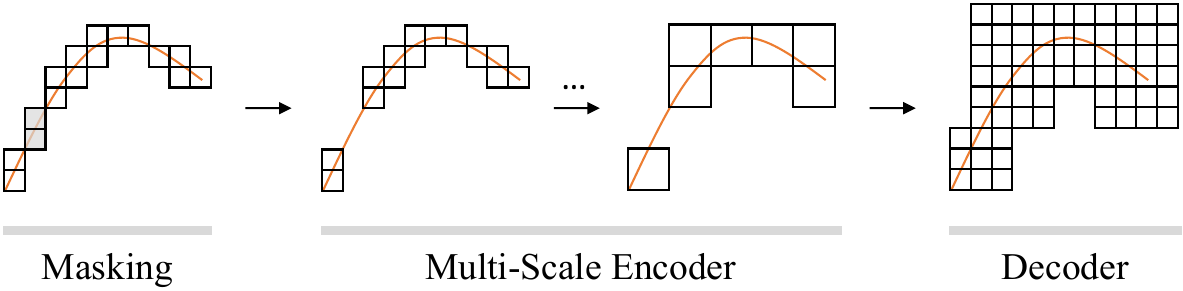}
	\vspace{-0.8em}
	\caption{Illustration of area expansion. The input point cloud (i.e., the orange curve) is voxelized and fed into the multi-scale encoder. The generative decoder can automatically expand visible features to potentially masked areas.}
	\label{fig:adapter_demo}
	\vspace{-1.5em}
\end{figure}

\paragraph{Transformer in Point Cloud Analysis.}
Transformer~\cite{ashish2017transformer} has become a competitive feature learning module in many computer vision tasks~\cite{carion2020detr,zhu2021deformabledetr}, which also inspires recent studies~\cite{liu2021groupfree,pan2021pointformer,yang20213dman,liu2021sparsepoint,yang2022eqpvrcnn,sheng2021ct3d,misra20213detr,mao2021votr,he2022voxset,deng2022vista,guan2022m3detr,liu20223dqueryis} for point cloud analysis.
Point Transformer~\cite{zhao2021pointtransformer} employs vector attention to better extract local features.
Point Transformer V2~\cite{wu2022pointtransv2} enhances Point Transformer and presents a more powerful and efficient model.
Fast Point Transformer~\cite{park2022fastpointtransformer} proposes a lightweight
self-attention layer and a voxel hashing-based architecture to boost computational efficiency.
Stratified Transformer~\cite{lai2022stratified} enlarges the effective receptive field at a low computational cost by sampling nearby points densely and distant points sparsely in a stratified way.
% CenterFormer~\cite{zhou2022centerformer} uses the feature of center candidates as queries and aggregates features from multi-scale feature maps through deformable attention~\cite{zhu2021deformabledetr}.
Object DGCNN~\cite{wang2021objectdgcnn} models 3D object detection as message passing on a dynamic graph and removes the necessity of non-maximum suppression.

\paragraph{Self-supervised Learning for Point Clouds.}
Point cloud representation learning without labels has been widely studied in recent years~\cite{liang2021gcc3d,ercelik2022sfb,li2022dpco,yin2022proposalcontrast,yu2022pointbert,afham2022crosspoint}.
OcCo~\cite{wang2021occo} occludes point clouds based on different viewpoints and learns to complete them.
PointContrast~\cite{xie2020pointcontrast} contrasts point-level features from two transformed views to learn discriminative 3D representations.
DepthContrast~\cite{zhang2021depthcontrast} learns features by considering voxels and
point clouds of the same 3D scene as data augmentations.
% CrossPoint~\cite{afham2022crosspoint} captures the correspondence between point clouds and rendered 2D images to constructively learn transferable point cloud representations.
4DContrast~\cite{chen20224dconstrast} leverages 4D signals in unsupervised pre-training to imbue 4D object priors into learned 3D representations.
% Point-BERT~\cite{yu2022pointbert} introduces a BERT-style pre-training strategy for 3D point clouds with a standard transformer network.
Inspired by the promising results achieved by MAE~\cite{he2022mae} in 2D vision, some works extend it into point clouds.
Point-MAE~\cite{pang2022pointmae} divides the point cloud into irregular point patches and aims to reconstruct the masked patches.
MaskPoint~\cite{liu2022maskpoint} represents the point cloud as discrete occupancy values and designs the decoder to discriminate masked real points and sampled fake points.
Differently, we explore MAE in the challenging outdoor point clouds, which have not yet been fully investigated.

%% file: section/method.tex
\section{Methodology}
In this section, we first review previous works in Sec.~\ref{subsec:preliminaries}.
Then, the designed sparse pyramid transformer and masked autoencoder are elaborated in Sec.~\ref{subsec:mae}.

\subsection{Preliminaries}
\label{subsec:preliminaries}
In contrast to conventional voxel-based detectors, pillar-based methods discretize the input point cloud with a grid of fixed size in the x-y plane, resulting in pillars rather than cubic voxels. This compact representation makes it achieve a good balance of efficiency and accuracy. In this section, we revisit the pillar-based representation and the extension of the sparse transformer on top of it.

\paragraph{Pillar-based Representation.}
PointPillars~\cite{lang2019pointpillar} is the pioneering pillar-based detector with 2D CNNs.
The 3D space is divided into equally distributed pillars which are voxels of infinite height.
The points are assigned to pillars to generate a feature vector.
Subsequently, the obtained pillar features are scattered back to their corresponding horizontal locations in the scene to form a dense 2D pseudo-image.
The pseudo-image is then processed by a feature pyramid network, which extracts multi-scale features using convolution layers with strides of 1$\times$, 2$\times$, and 4$\times$.
We refer the readers to \cite{lang2019pointpillar} for more details.

\paragraph{Sparse Transformer.}
SST~\cite{fan2022sst} is a transformer-based 3D detector operating on non-empty pillars.
Similar to Swin Transformer~\cite{liu2021swin}, SST divides the space into a list of non-overlapping windows with a fixed size. 
The self-attention is adopted among pillars within the same window.
Owing to its single-stride property, SST achieves impressive results for small object detection.

\paragraph{Analysis.}
Due to the self-occlusion of 3D objects, most of the points are sparsely distributed over the surface of the objects.
Spatial disconnection~\cite{chen2022lk3d} of sparse points can be exacerbated when extensive masking is applied.
For the visible points, it will be challenging to use a traditional convolution backbone like PointPillars to contain enough receptive fields.
To address this issue, inspired by SST, we introduce a simple yet effective transformer-based pyramid structure to achieve a large spatial scope.

\begin{figure*}[!t]
	\centering
	\includegraphics[width=1.9\columnwidth]{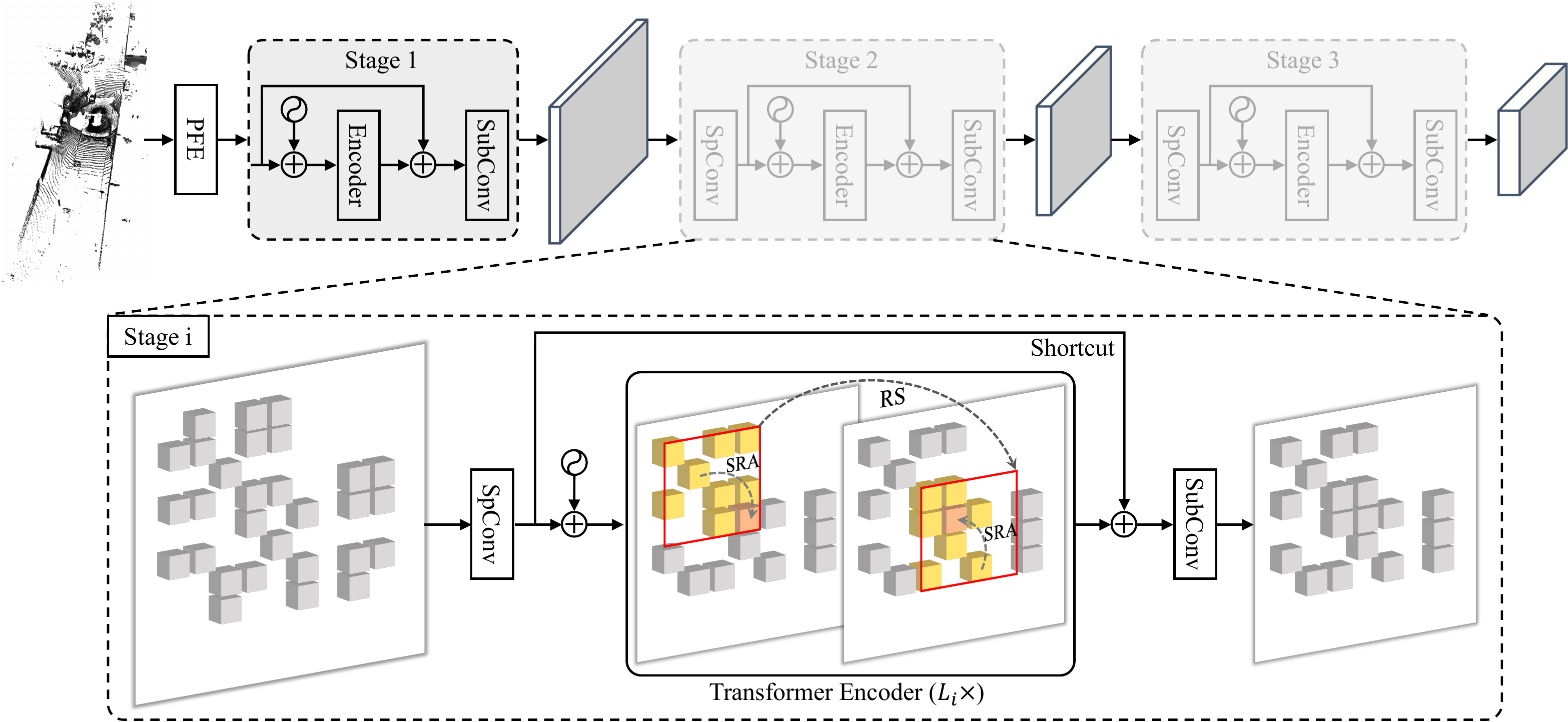}
	\vspace{-0.8em}
	\caption{Architecture of sparse pyramid transformer (SPT). The point clouds are fed into pillar feature encoding (PFE) to obtain a grid of pillars. The features and coordinates of the pillars constitute the tokens, which are then processed by three stages. The three stages have a similar structure, except that the second and third stages have a sparse convolution (SpConv) with a stride of 2 for downsampling. In each stage, there are $L_i$ transformer encoders and a shortcut connection, followed by a submanifold convolution (SubConv).}
	\label{fig:spt_stru}
	\vspace{-1.4em}
\end{figure*}

\subsection{Masked Autoencoder}
\label{subsec:mae}
\begin{figure*}[!t]
	\centering
	\includegraphics[width=1.9\columnwidth]{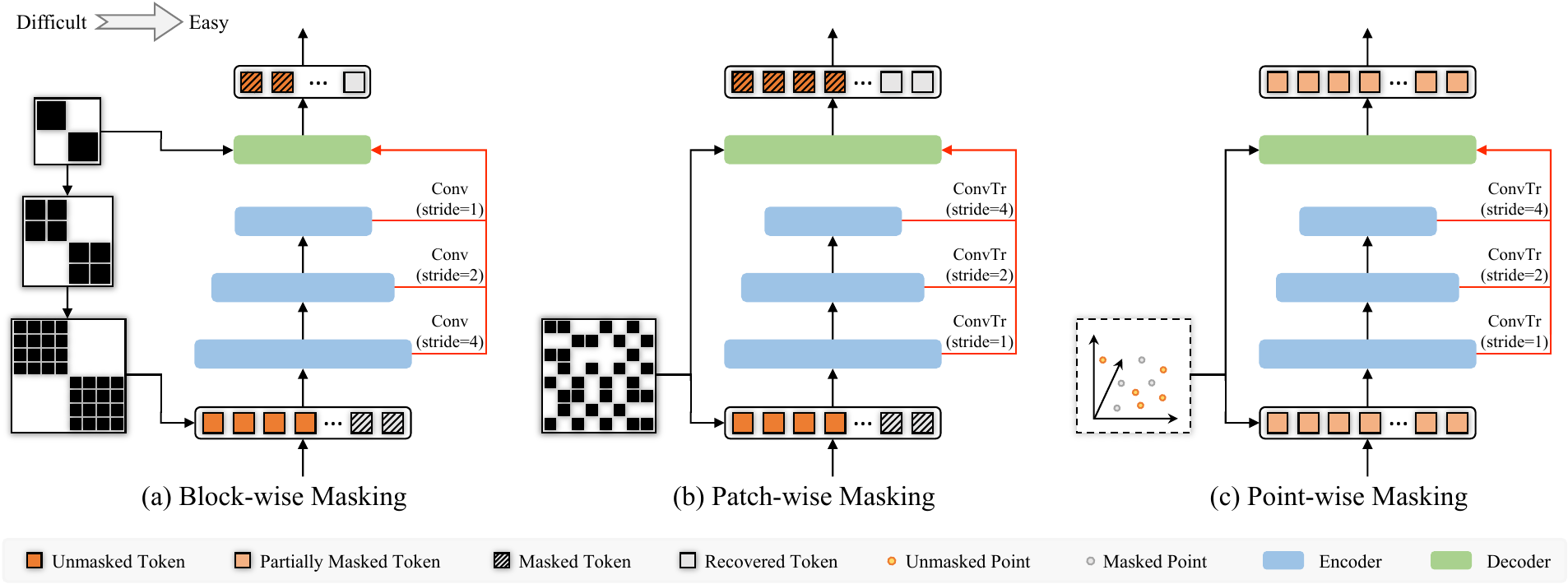}
	\vspace{-0.8em}
	\caption{Architecture of the masked autoencoder (MAE). The visible tokens are fed into the multi-scale encoder to create hierarchical features. Then, the generative decoder takes as input the multi-scale feature map and the masking map to unify the multi-scale features to a specific scale and recover the features of masked tokens. Finally, the recovered tokens are used for geometric reconstruction.}
	\label{fig:mae_stru}
	\vspace{-1.4em}
\end{figure*}

Inspired by the success of MAE~\cite{he2022mae} in 2D images, we develop the masked autoencoder for self-supervised learning on LiDAR point clouds, as shown in Figure~\ref{fig:mae_stru}.
The core idea is to use the encoder to create multi-scale representations from partial observations of the input.
The decoder is thereafter applied to unify the multi-scale features to a determined scale and expand the visible features to the underlying masked area.
Finally, the features of masked parts are processed by a head to reconstruct corresponding input point clouds.
After pre-training, the parameters of the encoder are used to warm up the backbone of the detection task.
Details are described below.

\paragraph{Multi-Scale Encoder.}
Unlike previous approaches~\cite{pang2022pointmae,liu2022maskpoint,hess2022voxelmae} that use a standard transformer encoder with a constant resolution for feature extraction, we exploit a hierarchical transformer architecture to better capture features from sparse LiDAR point clouds.
We present the overview of the Sparse Pyramid Transformer (SPT) in Figure~\ref{fig:spt_stru}.

Similar to PointPillars~\cite{lang2019pointpillar}, the input points $\mathcal{P}=\left\{p_i\right\}_{i=0}^{N-1}$ are converted to a grid of 2D pillars on bird's eye view by the pillar feature encoding (PFE) module.
Specifically, the pillar index of each point $p_i$ is first calculated as $v_i=(\lfloor\frac{x_i}{V_x}\rfloor, \lfloor\frac{y_i}{V_y}\rfloor)$, where $x_i$ and $y_i$ are coordinates of $p_i$ in the x-y plane, and $V_x$ and $V_y$ are the corresponding pillar size.
According to the pillar index, each point can be assigned to evenly divided pillar grids.
Since multiple points can potentially fall into the same pillar, a stack of PointNet~\cite{qi2017pointnet} is used to aggregate features from points to get pillar-wise features.
Finally, we take the pillars' features $\mathcal{F}\in\mathbb{R}^{M \times C}$ and the pillars' coordinates $\mathcal{C}\in\mathbb{R}^{M \times 2}$ as non-empty tokens (i.e., tokens involving at least one point).

The tokens are fed into three stages to generate feature maps of different scales.
In the first stage, non-empty tokens are taken as input and processed directly by a transformer encoder with a constant resolution, while in the other stages, tokens are first downsampled by a sparse convolution (SpConv) with a stride of $2$ and then passed through a encoder.

To construct the transformer encoder, we borrow the idea from recent works~\cite{fan2022sst,liu2021swin}.
In the stage $i$, it has $L_i$ encoder layers, each of which is composed of two sparse regional attention (SRA) and one region shift (RS).
To be specific, two SRA are applied to perform self-attention on the tokens that fall in the same region, accompanied by the positional embedding based on the positions of tokens in each region.
Between them, one RS is employed by adding offsets of half of the region size to expand the receptive field of the tokens to capture useful contexts.
The entire process used to update tokens' features can be formulated as:
\begin{equation}
\mathcal{F}=\text{SRA}\left(\text{SRA}\left(\mathcal{F}, \mathcal{C}\right), \text{RS}\left(\mathcal{C}\right)\right),
\end{equation}
where $\mathcal{F}$ is tokens' features, and $\mathcal{C}$ is tokens' coordinates.
After the transformer encoder, the perceptual fields of tokens are broadened, and the long-range contexts are aggregated. However, local features are still important to obtain local geometric details, especially for small objects such as pedestrians. Motivated by this, we add a shortcut to fuse previous features and then use a submanifold convolution (SubConv) to realize adaptive fusion.

\paragraph{Masking.}
Directly applying the original masking strategy~\cite{he2022mae,zhang2022pointm2ae,gao2022convmae,pang2022pointmae} to the last stage of the multi-scale encoder would make the pretext task too difficult, especially for small objects.
Because the masking granularity of the output tokens of the last stage is too large, making it hard to recover from unmasked parts.
To study the impact, we design three masking strategies with different granularities, which make the training task from difficult to easy.
We use the example of pedestrians to better understand the difficulty level of these three masking strategies.

\textit{Block-wise Masking} masks a portion of non-empty tokens from downsampled feature maps, e.g., stage 3, and tries to recover them, as shown in Figure~\ref{fig:mae_stru}(a).
Different from MAE~\cite{he2022mae}, the multi-scale structure requires the backtracing~\cite{zhang2022pointm2ae,gao2022convmae} to make the masked regions consistent across scales to avoid information leakage from previous stages.
Thus, inspired by \cite{zhang2022pointm2ae,gao2022convmae}, we first upsample the masking map to its original scale, then index the corresponding pixels according to the coordinates of the input tokens to determine whether they are masked or not, and finally feed only the unmasked tokens into the encoder.
In the case of a pedestrian, it can be seen that the whole body is masked out and needs to be recovered from the arms or elsewhere, making the task hard.
  
\textit{Patch-wise Masking} adopts a smaller masking granularity than block-wise masking by masking some of the upsampled tokens, as shown in Figure~\ref{fig:mae_stru}(b).
Since the granularity of the upsampled tokens is the same as that of the input tokens, masking consistency is naturally maintained.
It can be regarded as an easier task as only some parts of the pedestrian's body need to be restored.

\textit{Point-wise Masking} directly masks out a number of the input point clouds and reconstructs the masked points inside tokens, as shown in Figure~\ref{fig:mae_stru}(c).
In contrast to the two strategies discussed above, it is trivial to train a point-wise decoder to predict the coordinates of the masked points because the positional encoding would leak the information~\cite{liu2022maskpoint}.
Thus, we adopt a patch-wise decoder considering a patch as the smallest granularity, and each token needs to reconstruct the masked points inside the token.
It results in the simplest pre-training task since the entire structure of the body of a pedestrian is preserved, but some details require to be reconstructed.

\begin{figure}[!t]
	\centering
	\includegraphics[width=1.0\columnwidth]{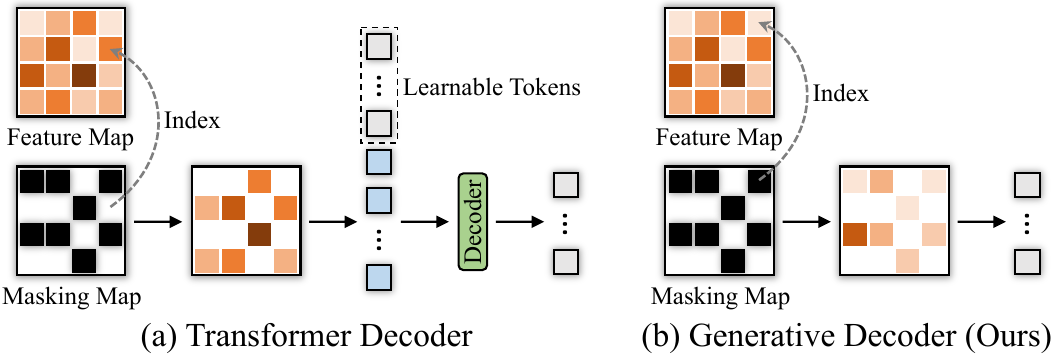}
	\vspace{-1.9em}
	\caption{Illustration of (a) the transformer decoder and (b) the proposed generative decoder. We show the invisible areas in black.}
	\label{fig:decoder_demo}
	\vspace{-1.4em}
\end{figure}

\paragraph{Generative Decoder.}
To enable varying masking granularity and unleash multi-scale representations for downstream tasks, we propose the Generative Decoder (GD) to fuse hierarchical features for reconstruction.
GD takes as input the visible tokens $E_1$, $E_2$, and $E_3$ from the multi-scale encoder to capture high-level semantic features and low-level geometric features, where $E_j$ denotes the tokens from stage $j$.
We then unify these tokens to the same scale, which is determined by the granularity of the corresponding masking strategy.
Specifically, we first transform the sparse tokens into a dense 2D feature map by scattering back tokens' features according to their corresponding coordinates and then performing a series of standard convolutions:
\begin{equation}
\label{eq:fuse}
D = \text{Conv}\left(\left[F_1(S(E_1)), F_2(S(E_2)), F_3(S(E_3))\right]\right),
\end{equation}
where $\left[\cdot,\cdot\right]$ is the concatenation function, $F$ is the convolution or the transposed convolution determined by the scale, $S$ is used to scatter back the sparse features, and \text{Conv} is a convolution with a kernel size of 3 for multi-scale feature fusion and area expansion.
Finally, we obtain the features $E_\text{mask}$ by indexing the feature vector on $D$ in terms of the coordinates of masked tokens, as shown in Figure~\ref{fig:decoder_demo}(b):
\begin{equation}
\label{eq:gather}
E = G\left(D\right),
\end{equation}
where $G$ is adopted to index features.

\begin{table*}[t]
    \centering
    \caption{Performance comparisons on the Waymo validation set. 0.2: using 20\% labeled data. $\dag$: we follow \cite{zheng2020ciassd,hu2022afdetv2,zhou2022centerformer,shi2022pillarnet} to use an extra IoU prediction head. $\ddag$: the refinement network of \cite{yang2022graphrcnn} is adopted to construct a two-stage detector. The results achieved by our \NickName are shown in bold, while top-performed results are shown in underline.}
    \vspace{-0.8em}
    \resizebox{1.9\columnwidth}!{
    \begin{tabular}{l|c|c|cc|cc|cc}
    \specialrule{1pt}{0pt}{1pt}
    \toprule
    \multirow{2}{*}{Methods} & \multirow{2}{*}{Voxel Size} &
    \multicolumn{1}{c|}{mAP/mAPH} &
    \multicolumn{2}{c|}{Vehicle 3D AP/APH} & \multicolumn{2}{c|}{Pedestrian 3D AP/APH} & \multicolumn{2}{c}{Cyclist 3D AP/APH} \\
     && L2 & L1 & L2 & L1 & L2 & L1 & L2 \\
    \midrule
    \bf{Two-stage:} & & & & &  \\
    % VoTr-TSD~\cite{mao2021votr} & [0.1, 0.1, 0.15] & -/- & 74.95/74.25 & 65.91/65.29 & -/- & -/- & -/- & -/-\\
    RSN ~\cite{sun2021rsn} & - & -/- & 75.10/74.60 &66.00/65.50 & 77.80/72.70 & 68.30/63.70 & -/- & -/- \\
    M3DETR~\cite{guan2022m3detr}& [0.1, 0.1, 0.15] & -/- & 75.71/75.08 & 66.58/66.02 & -/- & -/- & -/- & -/- \\
    Voxel RCNN~\cite{deng2021voxelrcnn} & [0.1, 0.1, 0.15] & -/- & 75.59/-  & 66.59/- & -/- & -/- & -/- & -/- \\
    Pyramid RCNN~\cite{mao2021pyramid}& [0.1, 0.1, 0.15] & -/- & 76.30/75.68 & 67.23/66.68 & -/- & -/- & -/- & -/-\\
    Part-A2-Net~\cite{shi2020part}& [0.1, 0.1, 0.15] & 66.92/63.84 & 77.05/76.51 & 68.47/67.97 & 75.24/66.87 & 66.18/58.62 & 68.60/67.36 & 66.13/64.93 \\
    PV-RCNN~\cite{shi2020pvrcnn} & [0.1, 0.1, 0.15] & 66.80/63.33 & 77.51/76.89 & 68.98/68.41 & 75.01/65.65 & 66.04/57.61 & 67.81/66.35 & 65.39/63.98 \\
    PV-RCNN++~\cite{shi2021pvrcnnplusplus}  & [0.1, 0.1, 0.15] & 71.66/69.45 & 79.25/78.78 & 70.61/70.18 & 81.83/76.28 & 73.17/68.00 & 73.72/72.66 & 71.21/70.19 \\
    % FSD~\cite{fan2022fsd} & [0.25, 0.25, 0.2] & 71.96/69.66 & 77.80/77.30 & 68.90/68.50 & 81.90/76.40 & 73.20/68.00 & 76.50/75.20 & 73.80/72.50 \\
    FSD~\cite{fan2022fsd} & [0.25, 0.25, 0.2] & 72.90/70.80 & 79.20/78.80 & 70.50/70.10 & 82.60/\underline{77.30} & 73.90/69.10 & 77.10/76.00 & 74.40/73.30 \\
    Graph R-CNN~\cite{yang2022graphrcnn} & [0.1, 0.1, 0.15] & 73.17/70.87 & \underline{80.77}/\underline{80.28} & \underline{72.55}/\underline{72.10} & 82.35/76.64 & 74.44/69.02 & 75.28/74.21 & 72.52/71.49 \\
    LiDAR-RCNN~\cite{li2021lidarrcnn}& [0.32, 0.32, 6] & 64.63/60.10 & 73.50/73.00 & 64.70/64.20 & 71.20/58.70 & 63.10/51.70 & 68.60/66.90 & 66.10/64.40\\
    SST\_TS~\cite{fan2022sst} & [0.32, 0.32, 6] & -/- & 76.22/75.79 & 68.04/67.64 & 81.39/74.05 & 72.82/65.93 & -/- & -/- \\
    % \midrule
    % \textbf{GD-MAE$^\ddag$ (Ours)} & \textbf{[0.32, 0.32, 6]} & \underline{\textbf{73.88}}/\underline{\textbf{71.34}} & \textbf{80.26}/\textbf{79.82} & \textbf{71.50}/\textbf{71.09} & \underline{\textbf{83.33}}/\underline{\textbf{76.90}} & \underline{\textbf{75.84}}/\underline{\textbf{69.68}} & \underline{\textbf{77.10}}/\underline{\textbf{76.03}} & \underline{\textbf{74.31}}/\underline{\textbf{73.27}} \\
    \textbf{GD-MAE$^\ddag$ (Ours)} & \textbf{[0.32, 0.32, 6]} & \underline{\textbf{74.11}}/\underline{\textbf{71.60}} & \textbf{80.21}/\textbf{79.78} & \textbf{72.37}/\textbf{71.96} & \underline{\textbf{83.10}}/\textbf{76.72} & \underline{\textbf{75.53}}/\underline{\textbf{69.43}} & \underline{\textbf{77.22}}/\underline{\textbf{76.18}} & \underline{\textbf{74.43}}/\underline{\textbf{73.42}} \\
    \midrule
    \textbf{One-stage:} & & & & & \\
    IA-SSD~\cite{zhang2022iassd} & - & 62.27/58.08 & 70.53/69.67 & 61.55/60.80 & 69.38/58.47 & 60.30/50.73 & 67.67/65.30 & 64.98/62.71 \\
    SECOND~\cite{yan2018second} & [0.1, 0.1, 0.15] & 60.97/57.23 & 72.27/71.69 & 63.85/63.33 & 68.70/58.18 & 60.72/51.31 & 60.62/59.28 & 58.34/57.05\\
    RangeDet~\cite{fan2021rangedet}& - & 64.96/63.20 & 72.90/72.30 & 64.00/63.60 & 75.90/71.90 & 67.60/63.90 & 65.70/64.40 & 63.30/62.10 \\
    CenterPoint-Voxel~\cite{yin2021center} & [0.1, 0.1, 0.15] & 68.25/65.81 & 74.78/74.24 & 66.66/66.17 & 75.95/69.75 & 68.42/62.67 & 72.27/71.12 & 69.69/68.59 \\
    Point2Seq~\cite{xue2022point2seq} & [0.1, 0.1, 0.15] & -/- & 77.52/77.03 & 68.80/68.36 & -/- & -/- & -/- & -/- \\
    AFDetV2~\cite{hu2022afdetv2} & [0.1, 0.1, 0.15] & 70.96/68.76 & 77.64/77.14 & 69.68/69.22 & 80.19/74.62 & 72.16/66.95 & 73.72/72.74 & 71.06/70.12 \\
    CenterFormer~\cite{zhou2022centerformer} & [0.1, 0.1, 0.15] & 71.20/68.93 & 75.20/74.70 & 70.20/69.70 & 78.60/73.00 & 73.60/68.30 & 72.30/71.30 & 69.80/68.80 \\
    PillarNet-34~\cite{shi2022pillarnet} & [0.1, 0.1, 6] & 70.97/68.43 & 79.09/78.59 & 70.92/70.46 & 80.59/74.01 & 72.28/66.17 & 72.29/71.21 & 69.72/68.67 \\
    MVF~\cite{zhou2019mvf} & [0.32, 0.32, 6] & -/- & 62.93/- & -/- & 65.33/- & -/- & -/- & -/-\\
    Pillar-OD~\cite{wang2020piilar-od} & [0.32, 0.32, 6] & -/- & 69.80/- & -/-  & 72.51/- & -/- & -/- & -/-\\
    PointPillars~\cite{lang2019pointpillar} & [0.32, 0.32, 6] & 62.61/57.57 & 71.56/70.99 & 63.05/62.54 & 70.60/56.69  & 62.85/50.24 & 64.35/62.26 & 61.94/59.93 \\
    CenterPoint-Pillar~\cite{yin2021center} & [0.32, 0.32, 6] & 65.98/62.21 & 73.37/72.86 & 65.09/64.62 & 75.35/65.11 & 67.61/58.25 & 67.76/66.22 & 65.25/63.77 \\
    SST~\cite{fan2022sst} & [0.32, 0.32, 6] & -/- & 74.22/73.77 & 65.47/65.07 & 78.71/69.55 & 70.02/61.67 & -/- & -/-\\
    VoxSeT~\cite{he2022voxset} & [0.32, 0.32, 6] & 69.13/66.22 & 74.50/74.03 & 65.99/65.56 & 80.03/72.42 & 72.45/65.39 & 71.56/70.29 & 68.95/67.73 \\
    \midrule
    \textbf{GD-MAE$_{0.2}$ (Ours)} & \textbf{[0.32, 0.32, 6]} & \textbf{70.24}/\textbf{67.14} & \textbf{76.24}/\textbf{75.74} & \textbf{67.67}/\textbf{67.22} & \textbf{80.50}/\textbf{72.29} & \textbf{73.18}/\textbf{65.50} & \textbf{72.63}/\textbf{71.42} & \textbf{69.87}/\textbf{68.71} \\
    \textbf{GD-MAE (Ours)} & \textbf{[0.32, 0.32, 6]} & \textbf{70.62}/\textbf{67.64} & \textbf{77.26}/\textbf{76.78} & \textbf{68.72}/\textbf{68.29} & \textbf{80.26}/\textbf{72.36} & \textbf{72.84}/\textbf{65.47} & \textbf{73.12}/\textbf{71.94} & \textbf{70.30}/\textbf{69.16}  \\
    \textbf{GD-MAE$^\dag$ (Ours)} & \textbf{[0.32, 0.32, 6]} & \textbf{72.90}/\textbf{70.43} & \textbf{79.40}/\textbf{78.94} & \textbf{70.91}/\textbf{70.49} & \textbf{82.20}/\textbf{75.85} & \textbf{74.82}/\textbf{68.79} & \textbf{75.75}/\textbf{74.77} & \textbf{72.98}/\textbf{72.03} \\
    \bottomrule
    \specialrule{1pt}{1pt}{0pt}
    \end{tabular}
    }
    \label{table:sota}
    \vspace{-1.4em}
\end{table*}

To set a baseline with the transformer decoder, we follow MAE~\cite{he2022mae} and some existing works~\cite{pang2022pointmae,zhang2022pointm2ae,he2022mae,gao2022convmae,hess2022voxelmae} to design the pipeline, as shown in Figure~\ref{fig:decoder_demo}(a).
For a fair comparison, we also adopt Eq.~\eqref{eq:fuse} to utilize multi-scale features for reconstruction, with the difference that the visible area is not expanded by setting the kernel size of the last convolution to 1.
Then, we follow Eq.~\eqref{eq:gather} to update the features of visible tokens according to the coordinates by indexing the feature map.
The transformer decoder accepts as input visible tokens and masked tokens with shared learnable embeddings.
Several flat transformer blocks~\cite{fan2022sst} are then applied to recover the features of masked tokens $E_\text{mask}$.

\paragraph{Reconstruction Target.}
For each masked token, the target is to recover the point cloud that falls within the corresponding token.
As different tokens contain a varying number of points, we randomly sample at most $K$ points as the target for reconstruction.
To stabilize the training, we normalize the point cloud to obtain $P_\text{mask} \in \mathbb{R}^{T \times K \times 3}$ by transforming it into local coordinates relative to the tokens, where $T$ is the number of masked tokens.
Given the features of masked tokens $E_\text{mask} \in \mathbb{R}^{T \times d}$, we project them using a linear function, followed by a reshape operation:
\begin{equation}
    \hat{P}_\text{mask}=\text{Reshape}(\text{Linear}(E_\text{mask})).
\end{equation}
Finally, the reconstruction loss is computed by l2 Chamfer Distance, which is formulated as:
\begin{equation}
	\mathcal{L}_\text{CD}=\text{ChamferDistance}(\hat{P}_\text{mask},P_\text{mask}).
\end{equation}

\begin{table}[!t]
	\centering
	\caption{Performance comparisons on the Waymo leaderboard.}
	\vspace{-0.8em}
        \setlength\tabcolsep{3pt}
	\resizebox{1.0\columnwidth}!{\begin{tabular}{l|c|cc|cc}
                \specialrule{1pt}{0pt}{1pt}
			\toprule
			\multirow{2}{*}{Methods} & \multicolumn{1}{c|}{mAP/mAPH} & \multicolumn{2}{c|}{Vehicle 3D AP/APH} & \multicolumn{2}{c}{Pedestrian 3D AP/APH} \\
			& L2 & L1 & L2 & L1 & L2 \\
			\midrule
              CenterPoint~\cite{yin2021center} & 72.20/69.10 & 80.20/79.70 & 72.20/71.80 & 78.30/72.10 & 72.20/66.40 \\
              PV-RCNN~\cite{shi2020pvrcnn} & 72.31/69.22 & 80.60/80.15 & 72.81/72.39 & 78.16/72.01 & 71.81/66.05 \\
              PV-RCNN++~\cite{shi2021pvrcnnplusplus} & 73.99/71.24 & 81.62/81.20 & 73.86/73.47 & 80.41/74.99 & 74.12/69.00 \\
			 FSD~\cite{fan2022fsd} & 75.17/72.66 & 82.70/82.33 & 74.40/74.06 & 82.90/\underline{77.88} & 75.93/71.26 \\
              Graph R-CNN~\cite{yang2022graphrcnn} & 75.82/73.05 & 83.55/83.12 & 
              \underline{76.04}/\underline{75.64} & 81.91/76.49 & 75.59/70.45 \\
              \midrule
			 \textbf{GD-MAE (Ours)} & \underline{\textbf{76.47}}/\underline{\textbf{73.37}} & \underline{\textbf{83.56}}/\underline{\textbf{83.16}} & \textbf{75.83}/\textbf{75.46} & \underline{\textbf{83.16}}/\textbf{77.05} & \underline{\textbf{77.10}}/\underline{\textbf{71.28}} \\
	   \bottomrule 
         \specialrule{1pt}{1pt}{0pt}
	\end{tabular}}
	\label{table:waymo_test}
        \vspace{-1.4em}
\end{table}

\begin{table*}[t]
    \centering
    \caption{Performance comparisons on the ONCE validation split. $\dag$: reproduced by us.}
    \vspace{-0.8em}
    \resizebox{2.1\columnwidth}!{
    \begin{tabular}{l|c|c|cccc|cccc|cccc}
    \specialrule{1pt}{0pt}{1pt}
    \toprule
    \multirow{2}{*}{Methods} & \multirow{2}{*}{Pre-trained} &
    \multirow{2}{*}{mAP} &
    \multicolumn{4}{c|}{Vehicle} & \multicolumn{4}{c|}{Pedestrian} & \multicolumn{4}{c}{Cyclist} \\
     &&& Overall & 0-30m & 30-50m & 50m-Inf & Overall & 0-30m & 30-50m & 50m-Inf & Overall & 0-30m & 30-50m & 50m-Inf \\
    \midrule
    % PointRCNN~\cite{shi2019pointrcnn} & \ding{55} & 28.74 & 52.09 & 75.45 & 40.89 & 16.81 & 4.28 & 6.17 & 2.40 & 0.91 & 29.84 & 46.03 & 20.94 & 5.46 \\
    % PointPillars~\cite{lang2019pointpillar} & \ding{55} & 44.34 & 68.57 & 80.86 & 62.07 & 47.04 & 17.63 & 19.74 & 15.15 & 10.23 & 46.81 & 58.33 & 40.32 & 25.86 \\
    PV-RCNN~\cite{shi2020pvrcnn} & \ding{55} & 53.55 & \underline{77.77} & \underline{89.39} & \underline{72.55} & \underline{58.64} & 23.50 & 25.61 & 22.84 & 17.27 & 59.37 & 71.66 & 52.58 & 36.17 \\
    IA-SSD~\cite{zhang2022iassd} & \ding{55} & 57.43 & 70.30 & 83.01 & 62.84 & 47.01 & 39.82 & 47.45 & 32.75 & 18.99 & 62.17 & 73.78 & 56.31 & 39.53 \\
    CenterPoint-Pillar$^\dag$~\cite{yin2021center} & \ding{55} & 59.07 & 74.10 & 85.23 & 69.22 & 53.14 & 40.94 & 48.43 & 34.72 & 20.09 & 62.17 & 73.70 & 56.05 & 40.19 \\
    CenterPoint-Voxel~\cite{yin2021center} & \ding{55} & 60.05 & 66.79 & 80.10 & 59.55 & 43.39 & \underline{49.90} & 56.24 & \underline{42.61} & \underline{26.27} & 63.45 & 74.28 & 57.94 & 41.48 \\
    \midrule
    SECOND~\cite{yan2018second} & \ding{55} & 51.89 & 71.19 & 84.04 & 63.02 & 47.25 & 26.44 & 29.33 & 24.05 & 18.05 & 58.04 & 69.96 & 52.43 & 34.61 \\
    w/ BYOL~\cite{grill2020byol} & \ding{51} & 51.63\down{0.26} & 71.32 & 83.59 & 64.89 & 50.27 & 25.02 & 27.06 & 22.96 & 17.04 & 58.56 & 70.18 & 52.74 & 36.32 \\
    w/ PointContrast~\cite{xie2020pointcontrast} & \ding{51} & 53.59\up{1.70} & 71.87 & 86.93 & 62.85 & 48.65 & 28.03 & 33.07 & 25.91 & 14.44 & 60.88 & 71.12 & 55.77 & 36.78 \\
    w/ DeepCluster~\cite{tian2017deepcluster} & \ding{51} & 53.72\up{1.83} & 72.89 & 83.52 & 67.09 & 50.38 & 30.32 & 34.76 & 26.43 & 18.33 & 57.94 & 69.18 & 52.42 & 34.36 \\
    \midrule
    \textbf{SPT (Ours)} & \ding{55} & \textbf{62.62} & \textbf{75.64} & \textbf{87.21} & \textbf{70.10} & \textbf{53.21} & \textbf{45.92} & \textbf{54.78} & \textbf{37.84} & \textbf{22.56} & \textbf{66.30} & \textbf{78.12} & \textbf{60.52} & \textbf{42.05} \\
    \textbf{w/ GD-MAE (Ours)} & \ding{51} & \underline{\textbf{64.92}}\up{\textbf{2.30}} & \textbf{76.79} & \textbf{88.01} & \textbf{71.70} & \textbf{55.60} & \textbf{48.84} & \underline{\textbf{58.70}} & \textbf{37.30} & \textbf{25.72} & \underline{\textbf{69.14}} & \underline{\textbf{80.29}} & \underline{\textbf{64.58}} & \underline{\textbf{45.14}} \\
    \bottomrule
    \specialrule{1pt}{1pt}{0pt}
    \end{tabular}
    }
    \label{table:once_val}
    \vspace{-1.6em}
\end{table*}

%% file: section/experiment.tex
\section{Experiments}
\subsection{Datasets}
\textbf{Waymo Open Dataset}~\cite{sun2020wod} is currently the largest dataset with LiDAR point clouds for autonomous driving.
There are total 798 training sequences and 202 validation sequences.
The evaluation protocol consists of the average precision (AP) and average precision weighted by heading (APH).
Also, it includes two difficulty levels: LEVEL\_1 denotes objects containing more than 5 points, and LEVEL\_2 denotes objects containing at least 1 point.
By default, we use a subset of the training splits by sampling every 5 frames from the training sequence for ablation studies.

\textbf{KITTI}~\cite{geiger2012kitti} includes 7481 LiDAR frames for training and 7518 LiDAR frames for testing.
As a common practice, the training data are divided into a \emph{train} set with 3712 samples and a \emph{val} set with 3769 samples.

\textbf{ONCE}~\cite{mao2021once} contains one million point clouds in total, in which 5k, 3k, and 8k point clouds are labeled as the training, validation, and testing split, respectively.
The remaining point clouds are kept unannotated, which are adopted by us for the pre-training.
The official evaluation metric is mean Average Precision (mAP), and the detection results are divided into 0-30m, 30-50m, and 50m-Inf.

\subsection{Implementation Details}
Our implementation is based on the codebase of OpenPCDet\footnote{\url{https://github.com/open-mmlab/OpenPCDet}}.
For the Waymo dataset, the detection ranges are set as $(-74.88, 74.88)$, $(-74.88, 74.88)$, and $(-2, 4)$, and the voxel size is $(0.32m, 0.32m, 6m)$.
For the KITTI dataset, the detection ranges are $(0, 69.12)$, $(-39.68, 39.68)$, and $(-3, 1)$, with a voxel size of $(0.32m, 0.32m, 4m)$.
For the ONCE dataset, the detection ranges are $(-74.88, 74.88)$, $(-74.88, 74.88)$, and $(-5, 3)$, and the voxel size is set to $(0.32m, 0.32m, 8m)$.
The pillar feature encoding module has two layers of MLPs with channel size of $[64, 128]$.
The 3D backbone consists of three stages, each of which has two transformer encoders with the input dimensions of $[128, 256, 256]$.
All of the transformer encoder layers have $8$ heads and inner MLP ratio of $2$.

During pre-training, we adopt several popular 3D data augmentation techniques: random flipping, scaling, and rotation.
For the masking, the mask ratio is set to $0.75$, and the $K$ is $64$ for point reconstruction.
The generative decoder first converts the dimensionality of the output feature map of the three stages to $128$ using three transposed convolutions and then transforms the concatenated features to $128$ using a convolution.
The model is trained for $30$ epochs with the AdamW~\cite{losh2019adamw} optimizer using the one-cycle policy, with a max learning rate of $3e^{-3}$.
During fine-tuning, in addition to the mentioned data augmentation, the copy-n-paste augmentation~\cite{yan2018second} is added to increase the number of training samples.
Similar to PointPillars, the multi-scale features are upsampled with transposed convolutions and then concatenated for the detection head.
We adopt a center-based head, and the training strategy and the target assignment strategy are the same as CenterPoint~\cite{yin2021center} in OpenPCDet.

\begin{table}[!t]
	\centering
	\caption{Ablation study on the Waymo validation set.}
	\vspace{-0.8em}
	\resizebox{1.0\columnwidth}!{\begin{tabular}{ccc|ccc}
			\toprule
			Pyramid & Shortcut & GD-MAE & Vehicle & Pedestrian & Cyclist \\
			\midrule
			 & & & 62.10/61.61 & 69.80/61.30 & 66.68/65.24 \\
			\ding{51} & & & 64.20/63.70 & 71.01/62.97 & 68.00/66.77 \\
			\ding{51} & \ding{51} & & 66.02/65.55 & 71.82/63.76 & 67.91/66.75 \\
			\ding{51} & \ding{51} & \ding{51} & \textbf{67.00}/\textbf{66.54} & \textbf{72.51}/\textbf{64.93} & \textbf{68.94}/\textbf{67.75} \\
			\bottomrule 
	\end{tabular}}
	\label{table:ablation_study}
        \vspace{-0.8em}
\end{table}

\begin{table}[!t]
	\centering
	\caption{Ablation study of the number of transformer encoders. Using 5\% data for training.}
	\vspace{-0.8em}
	\resizebox{0.8\columnwidth}!{\begin{tabular}{l|cccc|c}
			\toprule
			\multirow{2}{*}{\# Layer} & \multicolumn{4}{c|}{SST~\cite{fan2022sst}} & \multicolumn{1}{c}{Ours} \\
			& 4 & 6 & 8 & 10 & 6 \\
			\midrule
			 Vehicle & 57.25 & 56.56 & 56.71 & 56.49 &  \textbf{60.33} \\
			 Pedestrian & 55.95 & 54.18 & 53.58 & 53.96 & \textbf{57.28} \\
			\bottomrule 
	\end{tabular}}
	\label{table:abl_sst_depth}
	\vspace{-1.2em}
\end{table}

\subsection{Comparison with State-of-the-Art Methods}
We compare \NickName with other models on the Waymo dataset with a single frame LiDAR input.
As shown in Table~\ref{table:sota}, \NickName achieves new state-of-the-art results among all single-stage detectors on the Waymo validation set: it has 1.5 mAPH/L2 higher than the prior best single-stage model CenterFormer~\cite{zhou2022centerformer}.
Compared with the baseline, i.e., SST~\cite{fan2022sst}, \NickName improves the 3D APH at level 2 for vehicle and pedestrian by 5.42 and 7.12, respectively.
Compared with one-stage methods under the same settings of voxel size, \NickName outperforms them by a large margin (+4.93 APH/L2 for vehicle, +3.4 APH/L2 for pedestrian, and +4.3 APH/L2 for cyclist).
\NickName even outperforms PV-RCNN++~\cite{shi2021pvrcnnplusplus} by 0.98 mAPH/L2.
Equipped with a refinement network~\cite{yang2022graphrcnn}, \NickName surpasses all previously published methods.
Table~\ref{table:waymo_test} shows that \NickName also achieves impressive results on the Waymo leaderboard.

We evaluate the performance of the proposed \NickName on the ONCE validation split in Table~\ref{table:once_val}.
\NickName achieves significantly better detection results than previous strong detectors.
For example, the overall mAP of our approach is 64.92, which is 5.85 and 4.87 higher than the CenterPoint-Pillar and CenterPoint-Voxel, respectively.

\begin{table}[!t]
	\centering
	\caption{Ablation study of different scales of labeled data. Using 100\% data for pre-training.}
	\vspace{-0.8em}
	\resizebox{0.92\columnwidth}!{\begin{tabular}{c|cccc}
			\toprule
			w/ GD-MAE & 5\% & 10\% & 20\% & 100\% \\
			\midrule
			  & \textbf{59.97} & 63.58 & 65.35 & 67.34 \\
			 \ding{51} & \textbf{63.86}\up{\textbf{3.89}} & 65.62\up{\text{2.04}} & 67.14\up{\text{1.79}} & 67.64\up{\text{0.30}} \\
			\bottomrule
	\end{tabular}}
	\label{table:abl_label_ratio}
        \vspace{-0.8em}
\end{table}

\begin{table}[!t]
	\centering
	\caption{Ablation study of masked autoencoder on the KITTI \textit{val} set with moderate AP calculated by 40 recall positions.}
	\vspace{-0.8em}
	\resizebox{0.75\columnwidth}!{\begin{tabular}{c|ccc}
			\toprule
			w/ GD-MAE & Car & Pedestrian & Cyclist \\
			\midrule
			  & 81.46 & 46.52 & 65.59 \\
			 \ding{51} & \textbf{82.01}\up{\textbf{0.55}} & \textbf{48.40}\up{\textbf{1.88}} & \textbf{67.16}\up{\textbf{1.57}} \\
			\bottomrule
	\end{tabular}}
	\label{table:abl_mae_kitti}
	\vspace{-1.4em}
\end{table}

\subsection{Ablation Study}
\label{experiment:ablation}
In this section, we conduct a series of ablation experiments to comprehend the roles of different components.

\paragraph{Sparse Pyramid Transformer.}
The first and second rows of Table~\ref{table:ablation_study} demonstrate that the hierarchical architecture can consistently improve the detection accuracy of each category.
The impact on vehicle detection is greater than on other categories because vehicles are generally larger in size and therefore require a broader receptive field to obtain sufficient contextual information for 3D detection, such as size recognition.
The second and third rows of Table~\ref{table:ablation_study} show that a simple shortcut can further improve performance, suggesting that local geometric features are also important.
To further illustrate the necessity of the multi-scale structure, we scale up SST~\cite{fan2022sst} by deepening the network depth in Table~\ref{table:abl_sst_depth}.
We found that accuracy decreases with increasing depth, possibly due to overfitting.

\paragraph{Masked Autoencoder.}
The third and fourth rows of Table~\ref{table:ablation_study} show the effectiveness of the MAE, in which 20\% of the data is used for pre-training and fine-tuning.
It provides improvements of 0.99, 1.17, and 1.00 APH/L2 for vehicle, pedestrian, and cyclist, respectively.
The third column of Table~\ref{table:abl_label_ratio} shows that the performance can be further improved by 0.74 mAPH/L2 by using 100\% of the unlabeled data for pre-training.
To demonstrate the impact of pre-training on the data efficiency, Table~\ref{table:abl_label_ratio} shows the performance with different proportions of annotated data.
Without pre-training, the performance is increased by 1.99 mAPH/L2 when labeled data is added from 20\% to 100\%.
The gap is reduced to 0.2 mAPH/L2 when the pre-training is applied.
With only 5\% of annotated data available, MAE can significantly improve accuracy by 3.89 mAPH.
The impact of MAE is also verified on the ONCE and KITTI datasets in Tables~\ref{table:once_val} and \ref{table:abl_mae_kitti}, respectively, where consistent boosts can be observed.

\paragraph{Multi-Scale Encoders.}
We also verify the effect of \NickName on different multi-scale encoders such as convolution (CNN) and sparse convolution (SpCNN).
Table~\ref{table:abl_backbone} shows that \NickName consistently improves the accuracy of different multi-scale encoders. And we found that \NickName brings the highest gain for the proposed transformer encoder, which suggests that our simple decoder is particularly suitable for the transformer-based encoder.

\begin{table}[!t]
	\centering
        % \vspace{-0.8em}
	\caption{Ablation study of different multi-scale encoders.}
	\vspace{-0.8em}
	\resizebox{0.82\columnwidth}!{\begin{tabular}{c|cccc}
			\toprule
			w/ GD-MAE & CNN & SpCNN & Transformer \\
			\midrule
			  & 61.14 & 63.83 & 65.35 \\
			  \ding{51} & \textbf{61.48}\up{0.34} & \textbf{64.56}\up{0.73} & \textbf{66.40}\up{\textbf{1.05}} \\
			\bottomrule
	\end{tabular}}
	\label{table:abl_backbone}
        \vspace{-0.8em}
\end{table}

\begin{table}[!t]
	\centering
	\caption{Ablation study of different masking granularities. Using 5\% labeled data for fine-tuning.}
	\vspace{-0.8em}
	\resizebox{0.92\columnwidth}!{\begin{tabular}{l|c|cc}
			\toprule
			\multirow{2}{*}{Case} & \multirow{2}{*}{Block-wise~\cite{zhang2022pointm2ae,gao2022convmae}} & \multicolumn{2}{c}{Ours} \\
			& & Patch-wise & Point-wise \\
			\midrule
			 Vehicle & 61.60 & \textbf{62.65} & 62.41 \\
			 Pedestrian & 59.25 & \textbf{61.44} & 60.60 \\
			\bottomrule 
	\end{tabular}}
	\label{table:abl_mask_gra}
        \vspace{-1.6em}
\end{table}

\begin{table}[!b]
	\centering
	\vspace{-1.4em}
	\caption{Ablation study of the runtime of the decoder.}
	\vspace{-0.8em}
	\resizebox{0.5\columnwidth}!{\begin{tabular}{l|cc}
			\toprule
			Case & Baseline & Ours \\
			\midrule
			Runtime & 27.1ms & \textbf{3.2ms} \\
			\bottomrule 
	\end{tabular}}
	\label{table:decoder_time}
\end{table}

\paragraph{Masking Granularity.}
We study the effect of different masking granularities on LiDAR point clouds.
In Figure~\ref{fig:mae_stru}, for a fair comparison, all three masking strategies fuse hierarchical features to capture both high-level semantics and fine-grained patterns.
The only difference is the different masking granularity, which leads to different levels of difficulty in the pretext task.
From the last column of Table~\ref{table:abl_sst_depth} and the first column of Table~\ref{table:abl_mask_gra}, we can see that pre-training with block-wise masking can bring gains of 1.27APH/L2 and 1.97APH/L2 for the vehicle and pedestrian, respectively.
By using the finer masking granularity to ease the task, the performance is consistently improved, especially for pedestrians, as shown in the second and third columns of Table~\ref{table:abl_mask_gra}.
We can find that pre-training with the block-wise masking strategy brings less improvement than other methods, probably because the task is too difficult to learn effective features from sparse LiDAR point clouds.
And the overly simple task is also suboptimal for efficient feature learning, while an appropriate difficulty level, i.e., patch-wise masking, yields the best result.

\begin{table}[!t]
	\centering
	\caption{Ablation study of different decoders.}
	\vspace{-0.8em}
	\resizebox{0.8\columnwidth}!{\begin{tabular}{l|ccc|cc}
			\toprule
			\multirow{2}{*}{\# Decoder} & \multicolumn{3}{c|}{Baseline} & \multicolumn{2}{c}{Ours} \\
			& 1 & 2 & 3 & 1 & 2 \\
			\midrule
			 Vehicle & 66.24 & 66.20 & 66.23 & \textbf{66.54} & 66.40 \\
			 Pedestrian & 64.35 & 64.81 & 64.30 & \textbf{64.93} & 64.67 \\
			\bottomrule 
	\end{tabular}}
	\label{table:abl_decoder}
	\vspace{-1.8em}
\end{table}

\paragraph{Generative Decoder.}
In Table~\ref{table:abl_decoder}, we study the effect of different decoders.
The first three columns show the impact of different numbers of transformer decoders.
For pedestrians, the performance is better with two decoders.
If the number of decoders is increased or decreased, the results are adversely affected, indicating that performance is sensitive to the number of decoders.
And in our case, i.e., the generative decoder, we achieve the best results for both pedestrians and vehicles with only a convolution.
This shows that our framework could not only simplify MAE-style pre-training but also bring performance gains.
Table~\ref{table:decoder_time} demonstrates that the proposed decoder reduces runtime to about 0.12$\times$ compared to the transformer decoder.

% \subsection{Qualitative Results}
% Figure~\ref{fig:recon_vis} shows several examples of the reconstructed point clouds on the Waymo validation set.
% The model catches the distinctive LiDAR scans along the ground plane and demonstrates a knowledge of the basic geometry.

% \begin{figure}[!t]
% 	\centering
% 	\includegraphics[width=1.0\columnwidth]{figure/completion_vis.pdf}
% 	\vspace{-1em}
% 	\caption{Reconstruction results on the Waymo validation set. On the left is the visible input, in the middle is the result of the reconstruction and on the right is the ground truth.}
% 	\label{fig:recon_vis}
% 	\vspace{-1.8em}
% \end{figure}

%% file: section/conclusion.tex
\section{Conclusion}
In this paper, we present a much simpler paradigm dubbed \NickName for LiDAR point cloud pre-training in the MAE fashion.
We first propose the Sparse Pyramid Transformer as the multi-scale encoder to increase the spatial scope of the visible tokens.
Then, we introduce a Generative Decoder to simplify pre-training on hierarchical structures and enable fine-grained masking strategies for better feature learning.
Extensive experiments are conducted on Waymo, KITTI, and ONCE to verify the efficacy.
% \NickName sets new state-of-the-art detection results on the Waymo dataset.
% In the future, we will further explore MAE-style pre-training with a multi-frame rather than a single-frame for the continuous-time streams of LiDAR sensors on autonomous driving.

%% file: section/acknowledgment.tex
\vspace{-6pt}
\paragraph{Acknowledgement}
This work was supported in part by The National Nature Science Foundation of China (Grant Nos:  U1909203, 62273303), in part by Innovation Capability Support Program of Shaanxi (Program No. 2021TD-05), in part by the Key R\&D Program of Zhejiang Province, China (2023C01135).

%% file: section/appendix.tex
\section{Appendix}
This appendix contains the following sections:
Sec.~\ref{abl} reports more experimental results;
Sec.~\ref{vis} presents visualization results on the Waymo dataset.

\subsection{More experiments}
\label{abl}
\paragraph{Results on the KITTI Test Set.}
\NickName is compared with previous approaches using different 3D backbones, e.g., sparse convolutions (SpCNN) and Transformer, on the KITTI \textit{test} set.
As illustrated in Table~\ref{table:kitti_test}, \NickName achieves competitive results.

\begin{table}[!h]
	\centering
        \vspace{-0.8em}
	\caption{Performance comparisons on the KITTI \textit{test} set with AP calculated by 40 recall positions for the car class.}
	\vspace{-1.0em}
	\resizebox{0.85\columnwidth}!{\begin{tabular}{l|c|ccc}
        \specialrule{1pt}{0pt}{1pt}
		\toprule
		\multirow{2}{*}{Methods} & \multirow{2}{*}{Backbone} & \multicolumn{3}{c}{3D} \\
		& & Easy & Moderate & Hard \\
		\midrule
		 PointPillars~\cite{lang2019pointpillar} & CNN & 82.58 & 74.31 & 68.99 \\
		 SECOND~\cite{yan2018second} & SpCNN & 84.65 & 75.96 & 68.71 \\
		 VoTr-SSD~\cite{mao2021votr} & Transformer & 86.73 & 78.25 & 72.99 \\
          IA-SSD~\cite{zhang2022iassd} & PointNet++ & 88.34 & 80.13 & 75.04 \\
		 \midrule
		 \textbf{GD-MAE (Ours)} & \textbf{Transformer} & \textbf{88.14} & \textbf{79.03} & \textbf{73.55} \\
		\bottomrule
		\specialrule{1pt}{1pt}{0pt}
	\end{tabular}}
	\label{table:kitti_test}
	\vspace{-2.2em}
\end{table}

\paragraph{The Number of Points.}
As different tokens contain a varying number of points, we randomly sample at most $K$ points as the target for reconstruction.
Table~\ref{table:abl_k} shows the performance when different $K$ is adopted.

\begin{table}[!h]
	\centering
        \vspace{-0.8em}
	\caption{Ablation study of the number of the sampled points for the reconstruction target.}
	\vspace{-1.0em}
	\resizebox{0.52\columnwidth}!{\begin{tabular}{l|ccc}
			\toprule
			$K$ & 32 & 64 & 128 \\
			\midrule
			 Vehicle & \textbf{66.57} & 66.54 & 66.49 \\
			 Pedestrian & 64.64 & \textbf{64.93} & 64.67 \\
			\bottomrule
	\end{tabular}}
	\label{table:abl_k}
	\vspace{-2.2em}
\end{table}

\paragraph{Pre-training Epochs.}
Table~\ref{table:abl_epoch} shows the effect of the pre-training epochs.
We find that using more epochs can further improve performance, which demonstrates the learning capability of our model.
In the main text, all models are only pre-trained for 30 epochs to save training time.

\begin{table}[!h]
	\centering
        \vspace{-0.8em}
	\caption{Ablation study of the epoch for pre-training.}
	\vspace{-1.0em}
	\resizebox{0.6\columnwidth}!{\begin{tabular}{l|cccc}
			\toprule
			Epoch & 10 & 30 & 60 & 120\\
			\midrule
			 Vehicle & 66.23 & 66.54 & 66.82 & \textbf{66.89}\\
			 Pedestrian & 64.61 & 64.93 & 64.95 & \textbf{65.20} \\
			\bottomrule
	\end{tabular}}
	\label{table:abl_epoch}
        \vspace{-2.2em}
\end{table}

\paragraph{Different Pre-training Datasets.}
To prevent overfitting on the same dataset, we pre-train the model on the ONCE dataset and then fine-tune it on the Waymo dataset. As shown in the third row of Table~\ref{table:abl_ada}, GD-MAE consistently boosts the accuracy of all categories. 

\begin{table}[!h]
	\centering
        \vspace{-0.8em}
	\caption{Ablation study of different pre-training datasets.}
	\vspace{-1.0em}
	\resizebox{0.75\columnwidth}!{\begin{tabular}{c|c|ccc}
			\toprule
			w/ GD-MAE & Data. & Vehicle & Pedestrian & Cyclist \\
			\midrule
                 & - & 65.55 & 63.76 & 66.75 \\
                % Point-M2AE$^\dag$~[70] & Waymo & 66.15\up{0.60} & 64.17\up{0.41} & 67.18\up{0.43} \\
			  \ding{51} & Waymo & \textbf{66.54}\up{0.99} & \textbf{64.93}\up{\textbf{1.17}} & \textbf{67.75}\up{1.00} \\
			  \ding{51} & ONCE & \textbf{67.18}\up{\textbf{1.63}} & \textbf{64.82}\up{1.06} & \textbf{67.83}\up{\textbf{1.08}} \\
			\bottomrule
	\end{tabular}}
	\label{table:abl_ada}
        \vspace{-2.2em}
\end{table}

\paragraph{Masking Ratio.}
The impact of various masking ratios is displayed in Table~\ref{table:abl_mask_ratio}. We discover that a ratio of 75\% works best for creating a task that is adequately difficult for self-supervised pre-training.
If the masking ratio is too high, performance suffers dramatically.
The accuracy also degrades slightly with low making ratios.

\begin{table}[!h]
	\centering
        \vspace{-0.8em}
	\caption{Ablation study of different masking ratios. Using 5\% labeled data for fine-tuning.}
	\vspace{-1.0em}
	\resizebox{0.72\columnwidth}!{\begin{tabular}{l|ccccc}
			\toprule
			Ratio & 0.55 & 0.65 & 0.75 & 0.85 & 0.95 \\
			\midrule
			 Vehicle & 62.32 & 62.43 & \textbf{62.65} & 62.55 & 61.56 \\
			 Pedestrian & 60.91 & 61.19 & \textbf{61.44} & 60.97 & 60.19 \\
			\bottomrule 
	\end{tabular}}
	\label{table:abl_mask_ratio}
        \vspace{-1.8em}
\end{table}

\paragraph{The effect in multi-scale scenes.}
The vehicle category in the Waymo and ONCE datasets usually includes cars, buses, and trucks, which range from 4 to 12 meters in length.
The overall vehicle gain demonstrates the effectiveness of \NickName in multi-scale scenes.
In Table~\ref{table:abl_class}, we provide the results of subclasses of vehicle on the ONCE dataset.

\begin{table}[!h]
	\centering
        \vspace{-0.6em}
	\caption{Ablation study of different classes on the ONCE dataset.}
	\vspace{-1.0em}
	\resizebox{0.95 \columnwidth}!{\begin{tabular}{c|ccccc}
			\toprule
			w/ GD-MAE & Car & Bus & Truck & Pedestrian & Cyclist \\
			\midrule
			  & 76.94 & 59.31 & 34.45 & 45.92 & 66.30 \\
			  \ding{51} & \textbf{77.57}\up{\textbf{0.63}} & \textbf{67.08}\up{\textbf{7.77}} & \textbf{39.82}\up{\textbf{5.37}} & \textbf{48.84}\up{\textbf{2.92}} & \textbf{69.14}\up{\textbf{2.84}} \\
			\bottomrule
	\end{tabular}}
	\label{table:abl_class}
        \vspace{-1.2em}
\end{table}

\subsection{Qualitative Results}
\label{vis}
\begin{figure}[!h]
	\centering
	\includegraphics[width=0.98\columnwidth]{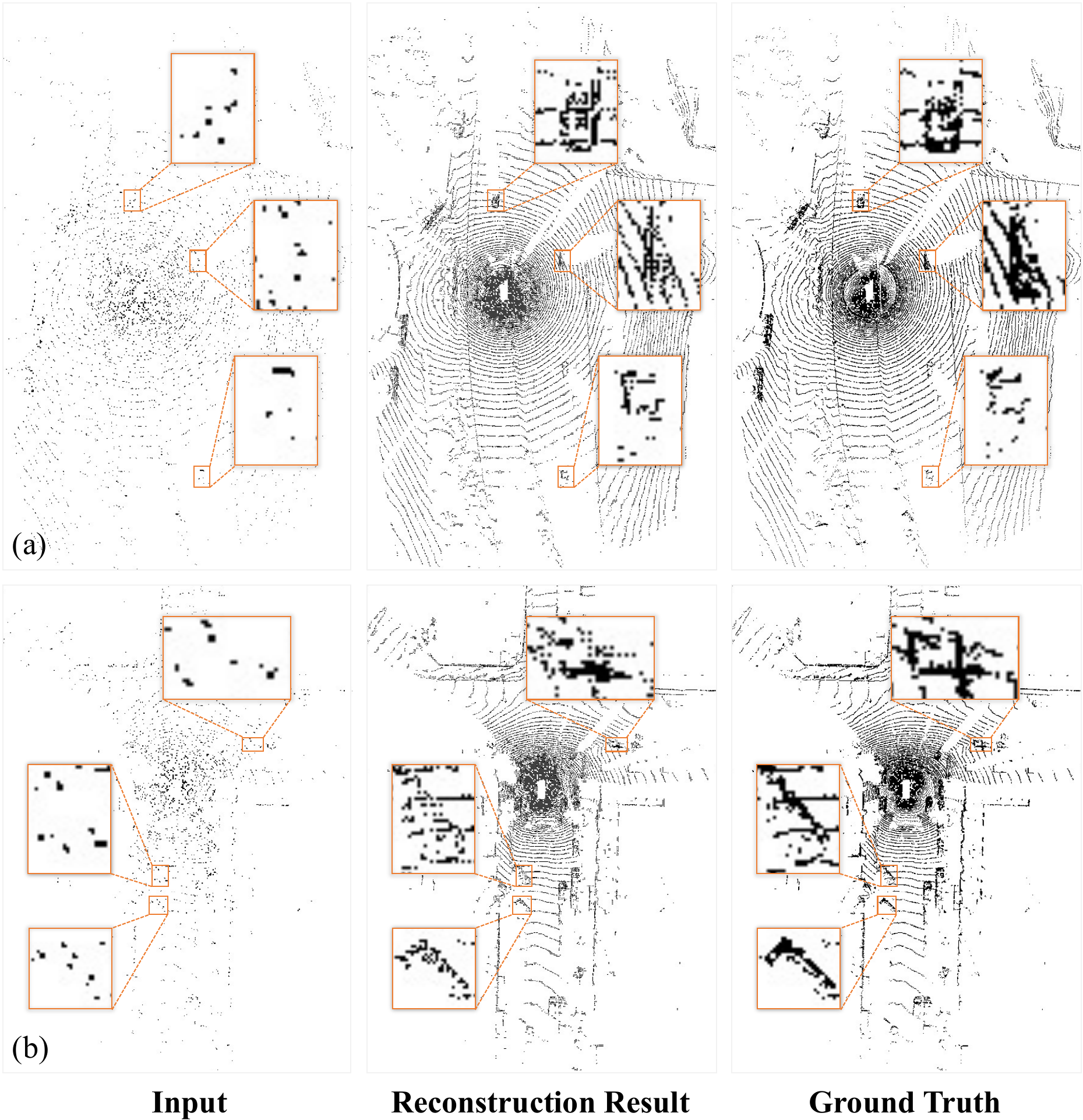}
	\vspace{-0.8em}
	\caption{Reconstruction results on the Waymo validation set. On the left is the visible input, in the middle is the result of the reconstruction and on the right is the ground truth.}
	\label{fig:recon_vis}
	% \vspace{-1.8em}
\end{figure}

\begin{figure*}[!t]
	\centering
	\includegraphics[width=2.0\columnwidth]{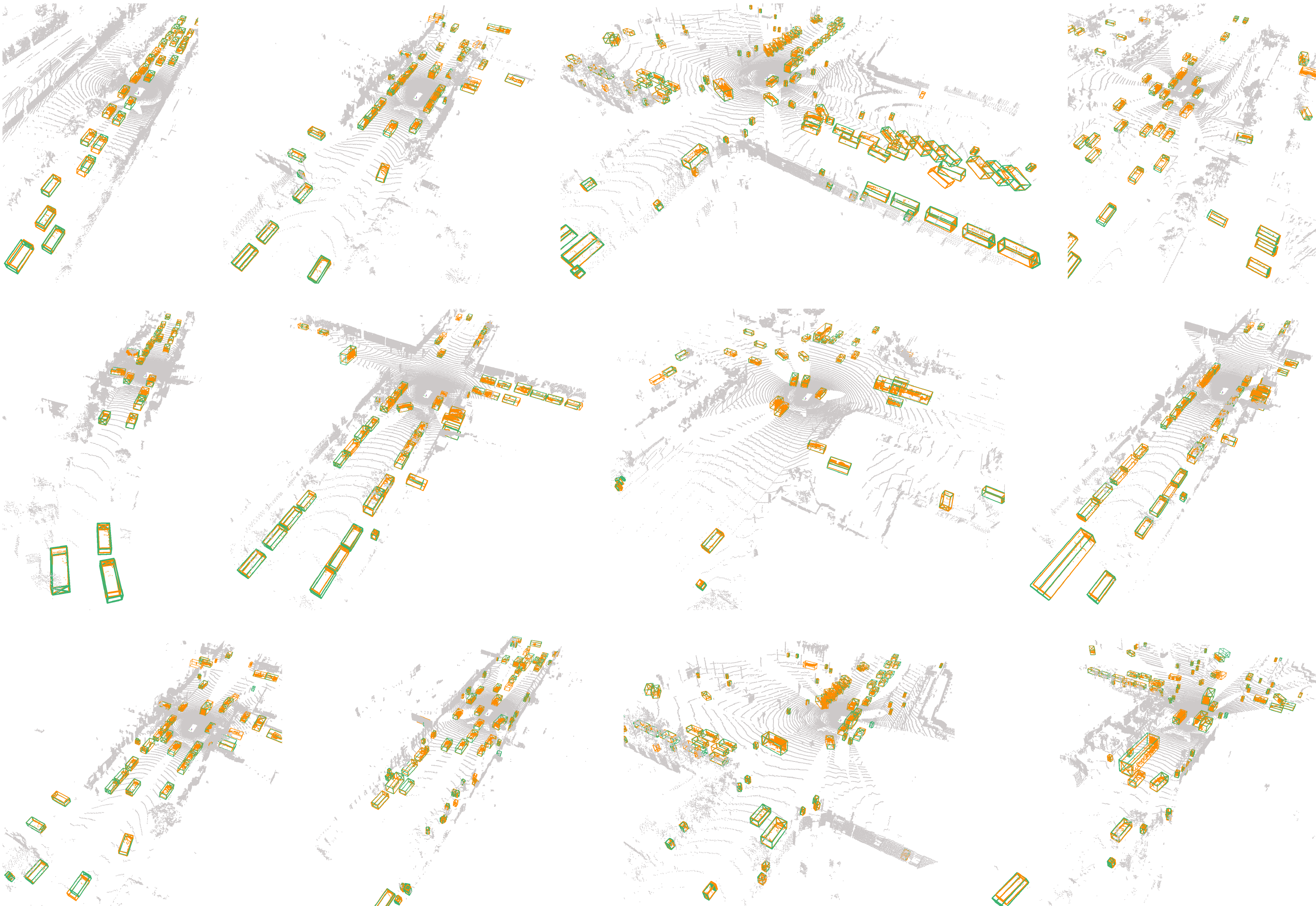}
	\vspace{-0.8em}
	\caption{Qualitative results of 3D object detection on the Waymo validation set. We show the raw point cloud in gray, points inside our detected bounding boxes in orange, ground truth in green bounding boxes, and our detected objects in orange bounding boxes.}
	\label{fig:det_vis}
\end{figure*}

Figure~\ref{fig:recon_vis} shows several examples of the reconstructed point clouds on the Waymo validation set.
The model catches the distinctive LiDAR scans along the ground plane and demonstrates a knowledge of the basic geometry.
Figure~\ref{fig:det_vis} illustrates the detection results of our method on the Waymo validation set.
Our model can predict accurate bounding boxes for distant and highly occluded objects, demonstrating the high-quality predictions of our model.

